\documentclass[letterpaper, 10 pt, conference]{ieeeconf}  % Comment this line out if you need a4paper
\makeatletter
\def\endfigure{\end@float}
\makeatother
\usepackage{amsmath,amsfonts}
\usepackage{algorithmic}
\usepackage{algorithm}
\usepackage{array}
\usepackage[caption=false,font=footnotesize,labelfont=rm,textfont=rm]{subfig}
\usepackage{textcomp}
\usepackage{stfloats}
\usepackage{url}
\usepackage{verbatim}
\usepackage{graphicx}
\usepackage{float}
\usepackage{bm}
\usepackage{multirow}
\usepackage{setspace}
\usepackage{threeparttable}
\usepackage{silence}
\usepackage{amssymb}
\usepackage{color}
\usepackage{xcolor}
\usepackage{pifont}
\usepackage{utfsym}
\usepackage{tabularx}
\usepackage{siunitx}
\usepackage[short]{optidef}
\usepackage{physics}
\usepackage[none]{hyphenat}
\usepackage[normalem]{ulem}
\UseRawInputEncoding
\usepackage{float}
\usepackage[hidelinks]{hyperref} 

\newcommand{\revM}[1]{\textcolor{black}{#1}}
\newcommand{\rM}[1]{\textcolor{black}{#1}}
\newcommand{\revB}[1]{\textcolor{black}{#1}}

\IEEEoverridecommandlockouts                              % This command is only needed if 
\title{\LARGE \bf
Safe Reinforcement Learning with a Predictive Safety Filter\\for Motion Planning and Control: A Drifting Vehicle Example
}

\author{Bei Zhou$^{1}$, 
Baha Zarrouki$^{2}$, Mattia Piccinini$^{2}$, Cheng Hu$^{1}$, Lei Xie$^{1*}$, Johannes Betz$^{2}$
\thanks{*Corresponding Author}% <-this % stops a space
\thanks{$^{1}$Bei Zhou, Cheng Hu, Lei Xie are with the State Key Laboratory of Industrial Control Technology, Zhejiang University, 310058 Hang zhou, China
	{\tt\small zhoubei@zju.edu.cn, 22032081@zju.edu.cn,  lxie@iipc.zju.edu.cn}}%
\thanks{$^{2}$Baha Zarrouki, Mattia Piccinini, and Johannes Betz are with the Professorship of Autonomous Vehicle Systems, Technical University of Munich, 85748 Garching, Germany; Munich Institute of Robotics and Machine Intelligence (MIRMI).
	{\tt\small  baha.zarrouki@tum.de, 
    mattia.piccinini@tum.de,
    johannes.betz@tum.de}}%
\thanks{*This work is supported by Jianbing Lingyan Foundation of Zhejiang Province, P.R. China (Grant No. 2023C01022) and Major Project of Science and Technology of Yunnan Province, China under Grant 202402AD080001.}
}

\begin{document}

\maketitle
\thispagestyle{empty}
\pagestyle{empty}

%%%%%%%%%%%%%%%%%%%%%%%%%%%%%%%%%%%%%%%%%%%%%%%%%%%%%%%%%%%%%%%%%%%%%%%%%%%%%%%%
\begin{abstract}
    Autonomous drifting is a complex and crucial maneuver for safety-critical scenarios like slippery roads and emergency collision avoidance, requiring precise motion planning and control.
    Traditional motion planning methods often struggle with the high instability and unpredictability of drifting, particularly when operating at high speeds. Recent learning-based approaches have attempted to tackle this issue but often rely on expert knowledge or have limited exploration capabilities. Additionally, they do not effectively address safety concerns during learning and deployment. To overcome these limitations, we propose a novel Safe Reinforcement Learning (RL)-based motion planner for autonomous drifting. Our approach integrates an RL agent with model-based drift dynamics to determine desired drift motion states, while incorporating a Predictive Safety Filter (PSF) that adjusts the agent's actions online to prevent unsafe states. This ensures safe and efficient learning, and stable drift operation. We validate the effectiveness of our method through simulations on a Matlab-Carsim platform, demonstrating significant improvements in drift performance, reduced tracking errors, and computational efficiency compared to traditional methods. This strategy promises to extend the capabilities of autonomous vehicles in safety-critical maneuvers.
	
\end{abstract}

%%%%%%%%%%%%%%%%%%%%%%%%%%%%%%%%%%%%%%%%%%%%%%%%%%%%%%%%%%%%%%%%%%%%%%%%%%%%%%%%
\section{INTRODUCTION}

The rising popularity of autonomous driving has promoted the \rM{development} of autonomous racing, a new research field that significantly \rM{extends} the limits of high-speed driving in dynamic environments \cite{AutonomousVehicles2022a,Piccinini2024}.
Drifting is a \rM{peculiar} racing technique used in motorsports, where drivers deliberately induce oversteering to generate more lateral driving force during sharp turns.
However, \rM{a successful} drifting requires maintaining \rM{vehicle control under unstable nonlinear dynamics}, with high sideslip angles and rear tire saturation. 
This makes drift motion planning a challenging task, due to the underactuated control problem and the strongly coupled longitudinal-lateral dynamics \cite{SimultaneousStabilization2016}.
Studying this aggressive driving behavior can extend the operational envelope of autonomous vehicles, enabling them to handle safety-critical situations like slippery roads and emergency obstacle avoidance.

\begin{figure}[t!]
    \centering
    \includegraphics[width=0.95\linewidth]{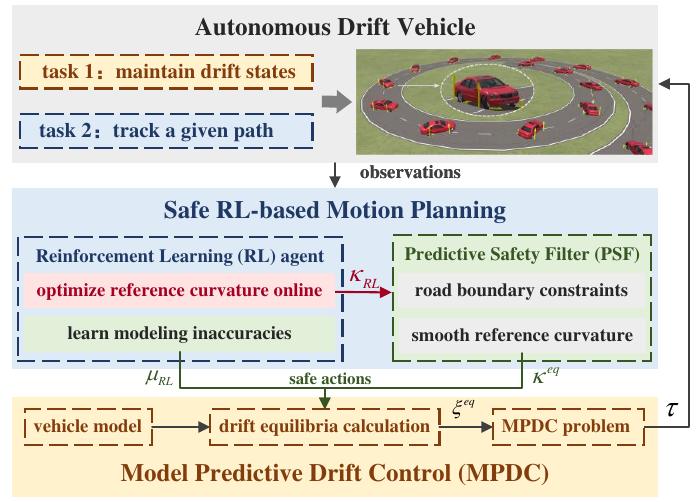}
    \caption{Overview of the proposed framework for autonomous drifting along a variable-curvature path. An RL-based motion planner learns the road friction and the local reference curvature to be tracked, which a predictive safety filter refines to ensure safe maneuvers. A model-based MPDC controller then generates low-level drift controls for the vehicle.
    }
    \label{intro}
\end{figure}

\revM{Most of the existing motion planning methods for autonomous drifting fall into two categories: model-based and learning-based.
Model-based planners use analytical vehicle models and techniques to plan the desired drifting states \cite{DynamicDrifting2023,ModelingControl2024a, AutonomousDrifting2024a}.}
However, their performance \revM{is limited by the} model fidelity, \revM{especially in non-nominal} environmental conditions.
\revM{Conversely, learning-based} approaches \revM{like} Reinforcement Learning (RL) and neural networks were used to learn drift equilibrium points \cite{TeachingVehicle2018} or optimal drift motions \cite{HighSpeedAutonomous2020a, DeepDrifting2022} \revM{without relying on analytical vehicle models}.
However, these data-driven methods often \revM{need expert knowledge and data to guide} the learning process.

\revM{Combining} model-based and \revM{learning-based} approaches \rM{has} become a new research direction in drift motion planning, where basic models serve as initial policies to bootstrap the data learning process \cite{AutonomousDrifting2016b}. 
This combination leverages the model-based component as a foundation and then learns residual model dynamics \cite{ djeumou2024one, LearningBasedMPC2022} or adjusts drift motions for performance improvement \cite{AutonomousDriving2022a, HarmonizedApproach2024a}.
However, the incorporation of data-driven components can compromise the \revM{system's stability and safety,} 
especially in the initial training phases.
To mitigate this risk, \revM{we introduce} a predictive safety filter into the drift \revM{motion planner} to ensure that the vehicle \rM{operates} within \revM{a safe domain.}

\subsection{Related Work}
\revB{
	%\revM{This section reviews the related work, first covering model-based drift motion planning strategies, and then discussing the challenges of model mismatches and how existing learning-based frameworks address them.}
	\revM{This section reviews the existing drift motion planners, from model-based to learning-based approaches.}
}

\subsubsection{Model-based Drift Motion Planning}
methods typically compute the drift states and vehicle trajectories for tracking by low-level controllers.
Chen et al. \cite{DynamicDrifting2023} used Model Predictive Control (MPC) to derive the desired drift states, %as a path tracking layer , 
but their kinematic tracking model could not accurately capture drifting behaviors due to coupled tire forces.
To address this, Weber et al. \cite{ModelingControl2024a} proposed a novel vehicle model incorporating weight transfer and wheel speed, and Lenzo et al. \cite{AutonomousDrifting2024a} exploited torque vectoring for autonomous drifting. However, their approaches can %aggravate the underactuated drift control problem and 
have limited generalization to parameter variations. Hu et al. \cite{NovelModel2024} used \revM{MPC for drift motion planning.
However, the MPC problem was computationally expensive, and they considered circular paths only.}
%However, these approaches can aggravate the underactuated drift control problem and have limited generalization due to parameter uncertainty.
%Besides, Zhao et al. \cite{HighSpeedCornering2024} first planned a Bezier curve, then used RL to iteratively minimize the cornering times.
%However, starting with pre-optimization and a minimum curvature principle may limit the exploration ability of the RL agent. \commM{Not clear: where are the pre-optimization and the minimum curvature principle in \cite{HighSpeedCornering2024}? Also, I'd mention \cite{HighSpeedCornering2024} in the next section, as it uses a learning-based method. Here, we'd need to mention other papers that used a model-based method: for example "Autonomous Drifting Using Torque Vectoring: Innovating Active Safety" or the other ones that you mention in the introduction.}

%In this paper, a model-free RL motion planner is adopted to learn the optimal tracking curvature directly from the given track. 
%This eliminates the need for expert drift-driving data and pre-trajectory optimization in the learning process. \commM{Don't say this here. Here, we need to only critically review the literature. Then, in Sec \ref{sec:contrib} we state the contributions.}

\subsubsection{Learning-based Drift Motion Planning}
approaches have been recently introduced to address the limitations of model-based drift motion planners, such as the capability to handle model mismatches, parameter uncertainties and environmental changes.
Neural networks were used in \cite{LearningBasedMPC2022} to learn a residual vehicle model, and in \cite{TeachingVehicle2018} to learn drift equilibrium points. 
However, the data-driven drift equilibria lacked interpretability and failed to ensure safe driving.
To address this, Cutler et al. \cite{AutonomousDrifting2016b} used model-based RL to learn optimal drift motions based on a classical vehicle model, while Djeumou et al. \cite{djeumou2024one} proposed a diffusion model for capturing complex trajectory distributions. Yet, capturing real-time environment variations and fitting them to the learned model remains challenging.
Additionally, \cite{HighSpeedAutonomous2020a} and \cite{DeepDrifting2022} adopted model-free RL to directly output drift control variables without explicit equations of motion, while Hou et al. \cite{AutonomousDriving2022a} and Zhao et al. \cite{HarmonizedApproach2024a} used RL to learn residual control variables.
However, these methods need large training datasets to generalize to unseen scenarios.
Zhao et al. \cite{HighSpeedCornering2024} used Bezier-based minimum-curvature optimization to guide an RL agent for optimal drift trajectories, improving learning efficiency but limiting exploration.
\subsubsection{Safe Reinforcement Learning (Safe RL)} incorporates safety constraints into RL to ensure that learned policies adhere to predefined safety standards during both training and deployment \cite{gu2024review}\cite{selim2022safe}. One common approach to enforce safety is through safety filters, which adjust the agent’s actions to prevent unsafe states. For example, Sootla et al. \cite{pmlr-v162-sootla22a} used state augmentation to maintain high probability of satisfying safety constraints, while Yang et al. \cite{yang2023safe} integrated probabilistic logic programming with policy gradient methods to enforce logical safety constraints in continuous action spaces. Additionally, Zarrouki et al. \cite{Safe-RL-Zarrouki} proposed a method combining safe RL with MPC, using a catalog of pre-optimized Bayesian optimization Pareto-optimal weight sets to constrain the RL action space.

\subsubsection{Predictive Safety Filters (PSFs)} ensure safety in dynamic systems, particularly when integrating learning-based controllers with safety-critical applications \cite{hsu2023safety}\cite{pmlr-v242-lavanakul24a}. Bejarano et al. \cite{Brunke} implemented action filtering, reward penalties, and safe resets that virtually eliminated safety violations in drone motion control applications. Wabersich et al. \cite{WABERSICH2021109597} developed a modular PSF framework, providing safety guarantees through an optimization-based approach, numerically demonstrated on a quadrotor. Leeman et al. \cite{pmlr-v211-leeman23a} further developed the concept using system level synthesis techniques to provide safety guarantees for any control policy, though limited to numerical validation.

\subsubsection{Critical Summary}

To our knowledge, the existing literature is limited by at least one of the following aspects:

\begin{itemize}
    \item Limited learning and adaptation capability: modeling errors can degrade the drift performance under changing environmental conditions \cite{DynamicDrifting2023, ModelingControl2024a, AutonomousDrifting2024a, NovelModel2024}.
    \item Dependence on prior expert knowledge and data in the learning process: drift driving data \revM{from professional drivers} \cite{TeachingVehicle2018, HighSpeedAutonomous2020a, AutonomousDriving2022a} or \revM{initial policies based on prior knowledge} \cite{AutonomousDrifting2016b, HarmonizedApproach2024a, HighSpeedCornering2024}.
    \item No examples of PSF within RL-based motion planners, to our current knowledge.
    \item MPC-based drift planners are computationally expensive \cite{NovelModel2024} and may not be suitable for real-time operation in rapidly-changing scenarios.
\end{itemize}
\subsection{Contributions} \label{sec:contrib}
\revM{To address these limitations, this paper makes the following key contributions:}
\begin{enumerate}
	\item We propose a Safe RL-based drift motion planner that learns the road friction coefficient and adapts the reference curvature of the path to be tracked, ensuring accurate drift equilibrium calculation for a low-level Model Predictive Drift Controller (MPDC).
	\item We design a Predictive Safety Filter (PSF) to enforce safe drifting maneuvers during training and inference while improving the RL agent performance.
	\item The proposed Safe RL motion planner requires no prior expert knowledge for training.
	\item Simulation results in Matlab-Carsim show that our Safe RL planner outperforms a state-of-the-art MPC planner benchmark in both closed-loop performance and computational efficiency for online operation.
\end{enumerate}	
\section{\rM{Framework Overview}}
Our framework (Fig. \ref{intro}) enables autonomous drifting along a variable-curvature path. We design a hierarchical planning and control architecture, where a Safe RL-based motion planner learns the local reference path curvature and road friction coefficient, while a Model Predictive Drift Controller (MPDC) generates low-level drift controls based on the RL planner's output. The RL planner adapts the tracking curvature and road friction to enhance closed-loop path tracking. We introduce a Predictive Safety Filter (PSF) to adjust the RL output curvature and ensure only safe actions reach the MPDC controller. The MPDC computes drift equilibrium points and generates controls to maintain the vehicle in the desired drift states (velocity, sideslip angle, yaw rate) while following the path.

\section{Model Predictive Drift Control} \label{sec:model}
This section introduces the model-based components of our framework, namely the vehicle dynamic model that we integrate into our Model Predictive Drift Control (MPDC). 

\subsection{Drift Vehicle Model}
We use the following single-track vehicle model \cite{ModelingControl2024a} to capture the drifting dynamics in our MPDC:

\setlength{\abovedisplayskip}{6pt} % Adjusts space before equations
\setlength{\belowdisplayskip}{6pt} % Adjusts space after equations
\begin{equation}
    \dot{V} = \frac{1}{m} (-F_{yf}\, \sin{(\delta-\beta)} + F_{yr}\,\sin{\beta} + F_{xr}\,\cos{\beta})
    \label{eq_single_track_V}
\end{equation}
\begin{equation}
    \dot{\beta} = \frac{1}{mV}(F_{yf}\, \cos{(\delta-\beta)} + F_{yr}\,\cos{\beta} - F_{xr}\,\sin{\beta}) - r
    \label{eq_single_track_beta}
\end{equation}
\begin{equation}
    \dot{r} = \frac{1}{I_z}(a \, F_{yf}\, \cos{\delta} - b\,F_{yr})
    \label{eq_single_track_r}
\end{equation}
where $V$ is the vehicle velocity, $\beta$ the sideslip angle, and $r$ the yaw rate, while the controls $\{\delta,F_{xr}\}$ are the steering angle at the front wheel and the longitudinal rear tire force. $m$ is the vehicle mass, $I_z$ is the yaw inertia, and $\{a,b\}$ are the distances from the center of mass to the front and rear axles. $F_{yf}$ and $F_{yr}$ are the front and rear lateral tire forces, which we describe with a Pacejka model \cite{TyreModelling1987}: 
\begin{align}
	F_{yi} &= -\mu F_{zi}\sin \big(C\,\arctan(B\,\alpha_{i})\big)  
	\label{Fyi}
\end{align}
where $\mu$ is the tire-ground friction coefficient, $F_{yi}$ and $ F_{zi}$ $(i = f, r)$ are the lateral and vertical tire forces.
$B$ and $C$ are tunable tire coefficients, and $\alpha_{i}$ $(i = f, r)$ are the tire sideslip angles.
When the vehicle performs drifting maneuvers, the rear tire forces will reach the friction limits, while the front tire forces remain below saturation to maintain lateral stability. 
In such conditions, the friction circle can be used to relate the lateral and longitudinal tire forces \cite{AutomatedVehicle2019}:
\begin{align} 
	F_{yr} &= \sqrt{(\mu F_{zr})^2 - F_{xr}^2 }
	\label{Fyr}
\end{align}
In this work, we advocate a RL agent to autonomously adjust the friction coefficient $\mu$ online (Section \ref{sec:RL}).
\subsection{Drift Equilibrium Points Calculation} \label{sec:equil_pts}
Drift equilibrium points are quasi steady-state solutions of the vehicle model \eqref{eq_single_track_V}-\eqref{eq_single_track_r} that describe a vehicle's drift state. A drift equilibrium point is defined as:
\begin{equation}
	\bm \xi^{eq} = [ V^{eq}, \beta^{eq}, r^{eq}, \delta^{eq},  F_{xr}^{eq} ]
	\label{eq: eq points}
\end{equation}
and we compute these points by setting the right-hand sides of the equations \eqref{eq_single_track_V}-\eqref{eq_single_track_r} to zero \cite{ControllerFramework2014a}, i.e., $\dot{V} = \dot{\beta} = \dot{r} = 0$. Since we have three equations and five unknowns, two additional relations are required to calculate the drift equilibrium states:
\begin{enumerate}    
    \item $\kappa^{eq} = r^{eq}/V^{eq}$, which relates the reference curvature $\kappa^{eq}$ to be tracked with the yaw rate and velocity. %In this work, we train an RL motion planner to optimize the curvature reference.
    \item $\delta^{eq}=\delta_0$, which assumes that the vehicle drifts with a constant steering angle along the reference path. %to facilitate deep drift states. 
\end{enumerate}
At each time step, our MPDC controller receives a curvature value from the motion planner and a constant steering angle as references, from which it computes the drift equilibrium and generates the drift controls. 
\subsection{MPDC Problem Formulation} \label{sec:MPC_control}
Our MPDC stabilizes the vehicle around the drift equilibrium points \eqref{eq: eq points} derived from the planner's outputs (Fig. \ref{intro}). 
The vehicle model \eqref{eq_single_track_V}-\eqref{eq_single_track_r} is then linearized around these equilibrium points \eqref{eq: eq points} and discretized using the forward-Euler method to formulate a quadratic program \cite{LearningBasedHierarchical2024}:
\begin{align} 
    \min_{\Delta \bm{\tau}_k} \quad & \sum_{k=1}^{N_p} \| \bm{\xi}_k - \bm{\xi}^{eq}_k \|^2_{\bm{Q}} + \sum_{k=1}^{N_c} \| \Delta \bm{\tau}_k \|^2_{\bm{R}} 
    \label{eq_MPC_control} \\
    \text{s.t.} \quad &
    \bm{\xi}_{k+1} = \bm f(\bm{\xi}_k, \Delta \bm{\tau}_k), \quad k \in \{1, \dots, N_p\} \notag \\
    & \bm{\tau}_k \in \mathcal{T}, \quad \Delta \bm{\tau}_k \in \mathcal{W}, \quad k \in \{1, \dots, N_c\} \notag
\end{align}
where $ \bm \xi = [ V, \beta, r, \delta,  F_{xr} ] $ are the states of the MPDC problem, while $\Delta \bm \tau = [\Delta \delta, \Delta F_{xr}]$ are the MPDC controls, representing the vehicle's input changes between consecutive time steps, and 
$\bm \xi^{eq}_k$ %= [ V^{eq}_k, \beta^{eq}_k, r^{eq}_k, \delta^{eq}_k, F_{xr_k}^{eq}]$ 
is the desired drift equilibrium point for the time step $k$. The first term of the cost function ensures stable drifting around $\bm \xi^{eq}_k$, while the second term ensures smooth control profiles.
Since drifting relies on precisely controlling the tire forces and counter-steering relative to the turning direction, having a reference for $\delta$ and $F_{xr}$ in $\bm \xi^{eq}_k$ enables the vehicle to quickly reach the desired drift states.
In \eqref{eq_MPC_control}, $N_p$ and $N_c$ are the predictive and control horizons, while $\mathbf{Q} $ and $\mathbf{R} $ are positive definite weighting matrices.
Finally, we constrain the control inputs to consider the actuation limits, with $\bm \tau \in \mathcal{T} :  \{|\delta| \leq \delta_{\text{th}}, F_{\text{min}} \leq F_{xr} \leq F_{\text{max}} \}$ and $\Delta \bm \tau \in  \mathcal{W}:  \{ |\Delta \delta| \le \Delta \delta_{\text{th}}, |\Delta F_{xr}| \le \Delta F_{\text{th}} \}$.

\section{\rM{Reinforcement Learning-based Motion Planning for Drifting}} 
\label{sec:RL}
Accurately calculating drift equilibria is crucial for controller performance. However, varying road conditions and modeling uncertainties pose significant challenges. To address this, we propose an autonomous learning approach using reinforcement learning (RL) to plan reference tracking curvature and adapt the road friction coefficient. Furthermore, we refine the RL agent’s learned curvature through a Predictive Safety Filter (PSF) (Section \ref{sec:safe}), ensuring that the generated actions remain within safe limits. This allows the policy to identify optimal drift equilibrium points online, enabling precise and adaptive control in dynamic environments while maintaining safety during both learning and deployment.

\subsection{Learning-based Motion Planning}
Our drift motion planner has two primary objectives: first, to determine the reference curvature at each time step, and second, to compensate for model inaccuracies by learning the local friction coefficient, thereby improving the accuracy of drift equilibrium calculations.

\subsubsection{Optimizing the Reference Curvature Online}
The path tracking problem for steady drift vehicles can be effectively addressed by decomposing the path into a series of circular arcs, each corresponding to a set of drift equilibrium points \cite{NovelModel2024}. The curvatures of these arcs play a critical role in both the vehicle’s path-following accuracy and its ability to maintain stable drifting performance.

Our RL agent derives the reference curvature $\kappa_{\mathrm{RL}}$ from the closest reference path curvature $\kappa_r$ to the vehicle's current position, by learning dynamic real-time adjustments $\varepsilon_{\kappa}$:
\begin{align} 
	\kappa_{\mathrm{RL}} = \kappa_r +  \varepsilon_{\kappa}
    \label{eq_curvat}
\end{align} 
The resulting curvature $\kappa_{\mathrm{RL}}$ from \eqref{eq_curvat} is then refined by our predictive safety filter (Section \ref{sec:safe}), which outputs a safe curvature $\kappa^{eq}$ to be used in the drift equilibrium calculation by the MPDC controller (Section \ref{sec:equil_pts}).

\subsubsection{Learning Modeling Inaccuracies} 
\label{sec:model error}

Capturing the complex drift dynamics with closed-form analytical models is challenging, and purely model-based calculations of drift equilibria often result in inaccuracies. One of the primary roles of our RL agent is to learn dynamic adjustments to the drift state calculations, compensating for such model discrepancies.

Model errors predominantly arise from the tire forces, as the tire model parameters are highly sensitive to varying road conditions and evolve over time due to wear. As equations (\ref{Fyi}) and (\ref{Fyr}) show, the tire forces are influenced by the road friction coefficient $\mu$, which, though difficult to measure directly, has a significant impact on vehicle dynamics.

To address this, we focus on $\mu$ as the key parameter for correcting discrepancies in the nominal drift model. Our RL agent returns a dynamically adapted $\mu_{\mathrm{RL}}$, starting from an initial guess $\mu_n$:
\begin{align}
\mu_{\mathrm{RL}} = \mu_n + \varepsilon_{\mu}
\label{eq_frict_coeff}
\end{align}
where $\varepsilon_{\mu}$ represents the dynamic adjustment learned by the RL agent. The adjusted $\mu_{\mathrm{RL}}$ is then used in the drift model \eqref{eq_single_track_V}-\eqref{eq_single_track_r} to refine the tire force calculations.

\subsection{Designing a Reinforcement Learning Motion Planner}
In this work, we employ the Deep Deterministic Policy Gradient (DDPG) algorithm \cite{ContinuousControl2019} for the RL agent. DDPG is an off-policy algorithm based on an actor-critic architecture, making it particularly well-suited for continuous action spaces. It leverages experience replay to improve sample efficiency, a crucial feature for enhancing the agent's learning process in dynamic environments.
\subsubsection{State Space}
The state space \( s \in \mathcal{S} \) of the RL agent is defined based on the key observations related to path tracking:
\begin{equation}
	\mathcal{S} = [e, \Delta \psi,  \delta, \kappa_r, e_{la}, h_e, h_{\mathrm{conv}}]
\end{equation}
where $e$ denotes the lateral path tracking error, %from the vehicle's current position to the closest point on the reference path, 
$\Delta \psi$ the yaw angle error, $\delta$ the steering angle, and $\kappa_r$ the curvature of the closest point on the given path.
Furthermore, $e_{la}$ denotes the predicted lateral path deviation projected at a look-ahead distance $x_{la}$ ahead of the vehicle. It is defined as $e_{la} = e + x_{la} \cdot\sin (\Delta{\psi})$, which allows the agent to predict future errors for corrective actions.
$h_e$ is a binary flag indicating whether the lateral error exceeds a predefined threshold \( e_{\text{th}} \), and is defined as:
\begin{align}
	h_e =
	\begin{cases}
		0, & \text{if} \,\,\, |e| < e_{\text{th}} \\
		1, & \text{otherwise}. 
	\end{cases}
\end{align}
Finally, the flag $h_{\mathrm{conv}}$ is set to 1 if the MPDC successfully converges to a solution for the given action; otherwise, it is set to 0.

\subsubsection{Action Space}
The continuous action space \( \mathcal{A} \) for the RL agent consists of two elements: 
\begin{equation}
	\mathcal{A} = %[\varepsilon_\kappa,  \varepsilon_\mu]
    [\kappa_{\mathrm{RL}}, \, \mu_{\mathrm{RL}}]
\end{equation}
where $\kappa_{\mathrm{RL}}$ and $\mu_{\mathrm{RL}}$ 
are the learned reference curvature \eqref{eq_curvat} and road friction coefficient \eqref{eq_frict_coeff}, respectively. 
Both actions are real-valued and constrained in specific ranges.

\subsubsection{Reward Design}
The reward function is designed to guide the RL agent toward optimal path tracking performance. It combines two components: 
\begin{align}
	r_f &=	-\arctan(|e| + \lambda|\Delta \psi|) \\
	r_p &=	-h_e \cdot |e| 
\end{align}
where $r_f$ encourages the agent to minimize both the lateral and yaw errors through the tunable parameter $\lambda$.
$r_p$ penalizes further the lateral error $e$ further when it exceeds a certain threshold (one lane width), i.e., when $h_e = 1$.
The overall reward $r$ is designed as:
\begin{align}
	r =
	\begin{cases}
		%	-\arctan(|e| + \lambda|\Delta \psi|) - h_e \cdot |e|, & h_f = 1 \\
		r_f + r_p, & \text{if} \,\,\, h_{\mathrm{conv}} = 1 \\
		-c, & \text{if} \,\,\, h_{\mathrm{conv}} = 0. 
	\end{cases}
    \label{eq_reward}
\end{align}
where \( c \) is a large constant that imposes a penalty if the MPDC fails to converge, signaling an infeasible action, i.e., if $h_{\mathrm{conv}} = 0$.

\subsubsection{Deep Neural Network Design}
The critic network of the DDPG agent consists of four hidden layers. 
The first two layers are fully connected layers with 256 and 128 neurons, respectively, which process the state and action inputs in parallel. 
The subsequent two hidden layers with 256 and 128 neurons are connected sequentially to integrate the outputs from the first two layers. 
All layers use the ReLU activations.

The actor network has four hidden layers: the first and second are fully connected layers with 256 neurons and ReLU activations.
The third and fourth layers are a tanh layer and a scale layer, which are utilized to adjust the output actions within the desired range.

\subsubsection{Training and Termination Conditions}

Training the RL agent involves episodes that begin with the vehicle at the starting point of a given clothoid-shaped path, moving at a constant initial speed \( V_0 \). At each time step, the RL agent provides the necessary curvature and road friction corrections to the low-level MPDC, which then computes the vehicle's controls. 
Each training episode concludes either when the maximum simulation time \( T_{\text{sim}} \) is reached, or when an early termination occurs if the lateral error exceeds a predefined threshold \( e_{\text{max}} \).

\section{Predictive Safety Filter for the RL Motion Planner}\label{sec:safe}
To ensure safety and feasibility of the motion planning system, we introduce a Predictive Safety Filter (PSF) \cite{PredictiveSafety2021a} into our reinforcement learning framework, as illustrated in Fig. \ref{intro}. We design the PSF to act as a supervisory layer that validates and refines the curvature references generated by the RL planner, preventing violations of track boundaries and ensuring dynamically feasible vehicle maneuvers.
The PSF is formulated as an optimal control problem (OCP), solved at each time step:
\begin{mini!}
	{\mathclap{\kappa \in \mathcal{K}}}{|| \kappa_{0}- \kappa_{\mathrm{RL}}||^2_{W_s} + \sum\limits_{i=0}^{N_s-1} ||\Delta \kappa_{i}||^2_{R_s} 
    \label{eq_PSF_cost_fun}}
	{\label{eq_PSF_problem}}{}
	\addConstraint{}{\bm{x}_{i+1} = \bm g(\bm{x}_{i}, \kappa_{i}) \label{eq_dyna_PSF}}
	\addConstraint{}{\bm{x}_{i} \in \mathcal{X},\ \kappa_{i} \in \mathcal{K},\ \Delta \kappa_{i} \in \mathcal{V}
    \label{eq_constr_PSF}}
\end{mini!}
where $\kappa_{\mathrm{RL}}$ is the reference curvature computed by the RL planner \eqref{eq_curvat}.
%The PSF ensures the vehicle drives within the track bounds in a future prediction horizon.
The predictive model $\bm g(\bm{x}, \kappa)$ in \eqref{eq_dyna_PSF} is the vehicle's path tracking dynamics\footnote{In \eqref{eq_track_err_dyna}, the quantities $V$, $\beta$, $r$ are kept constant over the PSF's prediction horizon, assuming small changes around the drift equilibria. This assumption enhances the PSF's computational efficiency without significantly affecting its performance.} \cite{AutomatedVehicle2019}, whose continuous-time formulation is:
\begin{equation}
	\begin{cases}
		\dot{e} = V \sin{\Delta \psi} \\ 
        \Delta \dot{\psi} = \dot\beta + r - \kappa \left( \frac{V \cos{\Delta \psi}}{1 - \kappa \, e} \right) 
	\end{cases}
    \label{eq_track_err_dyna}
\end{equation}
where $ \bm{x} = [e, \Delta \psi]$ are the PSF's states and the control is $\kappa$, which is the reference path curvature. Additionally, we introduce the safety constraints \eqref{eq_constr_PSF}, to ensure that the vehicle stays within the track bounds: $ \bm{x}_{i} \in \mathcal{X}: = \{ |e| < e_{\text{th}}, |\Delta \psi| < \Delta \psi_{\text{th}}\}$, constrain the planned curvature and its variation: $\kappa_{i} \in \mathcal{K}: = \{ \kappa_{\text{min}} < \kappa < \kappa_{\text{max}} \}$ and $\Delta \kappa_{i} \in \mathcal{V}: = \{|\Delta \kappa| < \Delta \kappa_{\text{th}} \}$, $\forall i \in [0,1,\dots,N_s-1]$.

The first term of the cost function \eqref{eq_PSF_cost_fun} ensures that the PSF's output curvature $\kappa_0$ at the first time step is close to the RL plan $\kappa_{\mathrm{RL}}$, while satisfying the safety constraints \eqref{eq_constr_PSF}. The second term in \eqref{eq_PSF_cost_fun} penalizes the curvature variations over the PSF's horizon, with $W_s$ and $R_s$ being tunable weights. The prediction horizon of the OCP \eqref{eq_PSF_problem} contains $N_s$ points.

The PSF's reference curvature for the current time step $\kappa_0$ is used to compute the corresponding drift equilibrium point (Section \ref{sec:equil_pts}), which the MPDC drift controller then tracks (Section \ref{sec:MPC_control}).

Algorithm \ref{alg1} overviews the training of our safe RL motion planner.
\begin{algorithm}[ht]
	\caption{Training our Safe RL Motion Planner} 
	\label{alg:alg1}
	\setstretch{1.1}
	\footnotesize
	\begin{algorithmic}
		\STATE
		Initialize RL\_agent \\
		{\textbf{for}} episode $n = 1 \ \textbf{to} \  N$ \textbf{do}\\
		\hspace{0.3cm} 	Initialize the vehicle with $V = V_0$ at the start of the training path \\
		\hspace{0.3cm} $\bm s_0 = [e, \Delta \psi, \delta, \kappa_r, e_{la}, h_e, h_{\mathrm{conv}}] \gets$ get initial observations \\
		
		\hspace{0.3cm} {\textbf{for}} time $t = 0 \ \textbf{to} \ T_{\text{sim}}$ \textbf{do}\\
		\hspace{0.6cm} $\bm a_t = [\kappa_{\mathrm{RL}}, \mu_{\mathrm{RL}}] = \text{RL\_agent}(\bm s_t) \gets$ compute RL actions \\	
		\hspace{0.6cm} $\kappa_0 = \text{PSF}(\kappa_{\mathrm{RL}}) \gets$ compute safe curvature \\
		\hspace{0.6cm} $\bm \xi^{eq}_t = \text{drift\_equil}(\kappa_0, \mu_{\mathrm{RL}}) \gets$ compute drift equilibrium \\ 
		\hspace{0.6cm} $\bm \tau_t = \text{MPDC}(\bm \xi^{eq}_t) \gets$ compute vehicle controls \\
		\hspace{0.6cm} $\bm s_{t+1} = \text{vehicle}(\bm s_{t}, \bm \tau_t) \gets$ get next state \\
		\hspace{0.6cm} $r_{t+1} = \text{reward}(\bm s_{t+1}) \gets$ compute reward \\
		\hspace{0.6cm} RL\_agent.\text{update}($\bm s_t, \bm a_t, r_t, \bm s_{t+1}$) $\gets$ train RL agent \\
		\hspace{0.6cm} \textbf{if} $|e| > e_{\text{max}}$ \textbf{then} break $\gets$ early termination \\

		\hspace{0.3cm} {\textbf{end for}} \\
		{\textbf{end for}}
		
	\end{algorithmic}
	\label{alg1}
\end{algorithm}
\section{Simulation Results} \label{sec:sim}
\subsection{Simulation Setup}
We train the RL agent on a laptop with an AMD Ryzen 7 5800H CPU (3.20 GHz) and 16 GB of RAM, using MATLAB 2022b. The drift vehicle model is from the Carsim 2019 simulator. 

The MPDC controller's discretization step matches the simulation sampling time ($T_\text{MPDC} = T_{\text{sim}} = 0.05$ s), while the RL motion planner operates with a longer sampling time of $T_\text{RL,planner} = 0.1$ s to reduce the computational load. Each training episode lasts $T_\text{sim} = 18$ s, corresponding to $180$ RL agent steps. 

The RL agent is trained over a maximum of $N = 3000$ episodes, with a learning rate $\alpha = 0.001$ and a discount factor $\gamma = 0.99$. The target networks update at 10 Hz, with a gradient step of 10. During training, mini-batches of size $m = 512$ are randomly sampled from an experience buffer with a capacity of $R = 20000$. The action noise has a standard deviation $\omega_n = 0.2$, and decays at a rate $\sigma_n = 5 \cdot 10^{-6}$. The main system parameters are summarized in Table \ref{parameters}, with additional drift vehicle details in \cite{ZHOU2025104941}.

\begin{table}[!ht]
\caption{Main parameters of our framework.}
\label{parameters}
\centering
\resizebox{0.8\linewidth}{!}{
\begin{tabular}{ll|ll}
	%		{\hsize}{@{}@{\extracolsep{\fill}}ll| ll@{}}
	\hline
	Parameters  & Value & Parameters  & Value  \\
	\hline
	$e_{\text{max}}$ & 5 m	 					& $N_s$ & 10 \\
	$e_{\text{th}}$ & 1.5 m	 					& $W_s$ & 10 \\
	$\Delta \psi_{\text{th}}$ & 0.2 rad 		& $R_s$ & 1 \\		
	$\kappa_{\text{th}}$ & 0.05	m$^{-1}$				&  $\mu_{\text{th}}$ & 0.15 \\
	$\Delta\kappa_{\text{th}}$ & 0.01 	m$^{-1}$	& $\kappa_\text{min}$ & 0.01	m$^{-1}$	\\
	$\kappa_\text{max}$ & 0.1 	m$^{-1}$						& $\delta_0$	& -0.5 rad	\\
	$V_0$	& 16.67 m/s 			              & $\lambda$ & 10\\
                                 $c$ & 100  \\
	\hline
\end{tabular}}
\end{table}

\subsection{Benchmarks and Ablation Studies}
To evaluate the effectiveness of the proposed safe RL-based planner, we compare it against one benchmark and two ablation scenarios. The benchmark is an MPC planner from \cite{NovelModel2024}, which we enhance by using a variable-curvature path prediction instead of a circular path. The ablation studies are as follows:

\begin{itemize}
    \item \textbf{RL Planner (only $\kappa$)}: Our proposed RL planner that learns only the reference curvature $\kappa$, without the PSF.
    \item \textbf{RL Planner ($\kappa$ and $\mu$)}: Our RL planner that learns both the curvature $\kappa$ and road friction coefficient $\mu$, without the PSF.
    \item \textbf{Safe RL Planner ($\kappa$ and $\mu$ + PSF)}: Our full RL planner, which learns both $\kappa$ and $\mu$, and incorporates the PSF for safety-enhanced motion planning.
\end{itemize}

In all cases, the low-level MPDC drift controller is unchanged, ensuring that the differences in performance are solely attributed to the changes in the planner configurations.

\subsection{Training Results}
\begin{figure}[h!]
\centering
\includegraphics[width=0.9\linewidth]{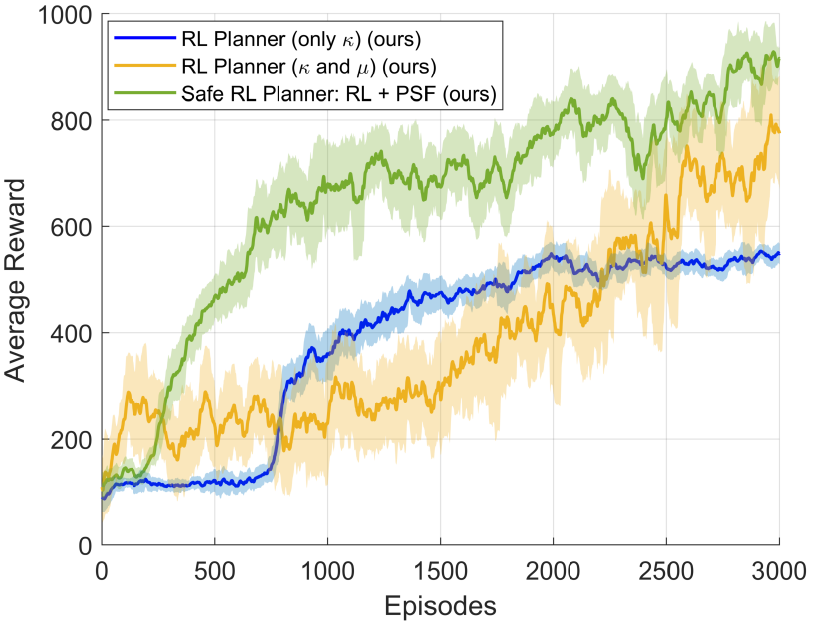}
\caption{The average rewards, smoothened over 50 episodes, show that the safe RL planner outperforms the two ablations: the PSF enhances training efficiency and boosts performance.}
\label{fig:training}
\end{figure}

The training process for the proposed RL planners is shown in Fig. \ref{fig:training}. Additionally, the tracking performance of all planners is presented in Fig. \ref{final_tracking} and Fig. \ref{fig:box_plots}, while the drifting performance is summarized in Table \ref{tab:drifting}. In the following, we provide a detailed analysis of the simulation results.

\subsection{Effect of Learning the Reference Curvature}
\begin{figure}[!ht]
	\centering
	\subfloat[Box plot for lateral path deviations.]
    {\includegraphics[width=1\linewidth]{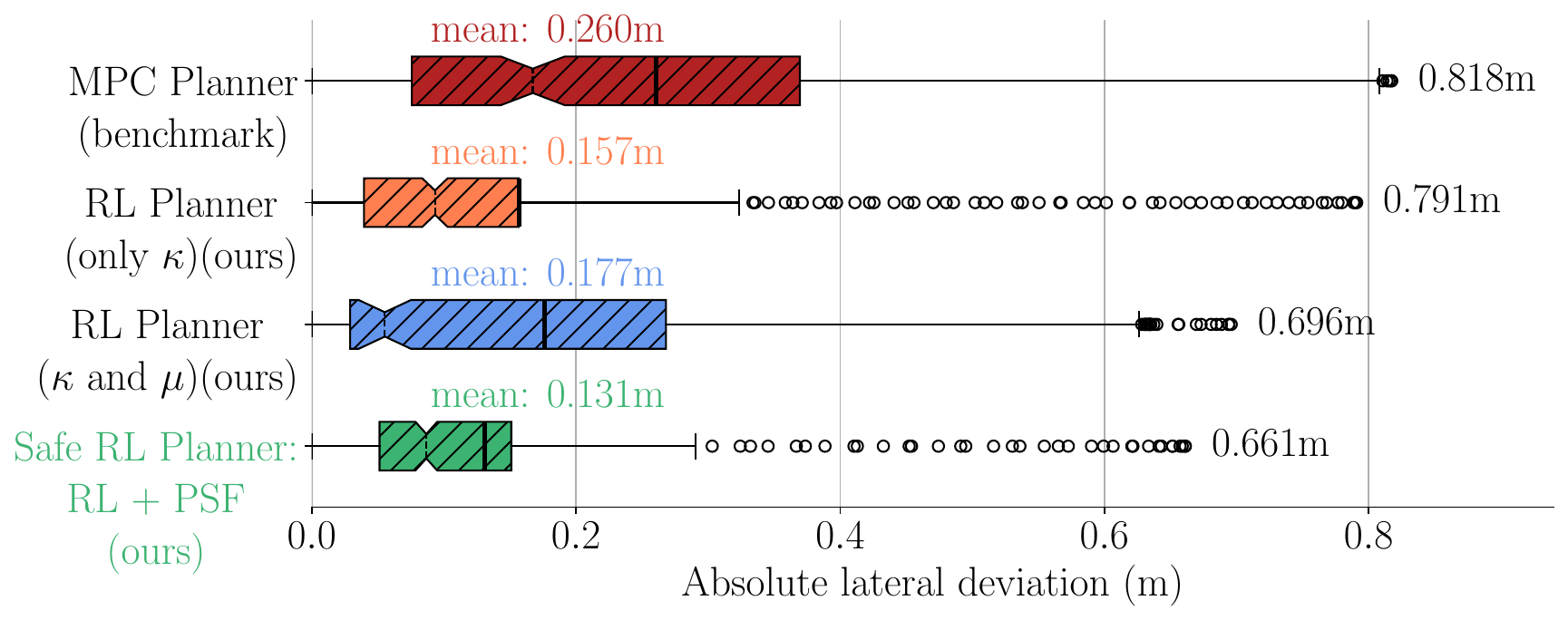}%
	\label{fig:box_e}}
	\hfil
	\subfloat[Box plot for heading deviations.]{\includegraphics[width=1\linewidth]{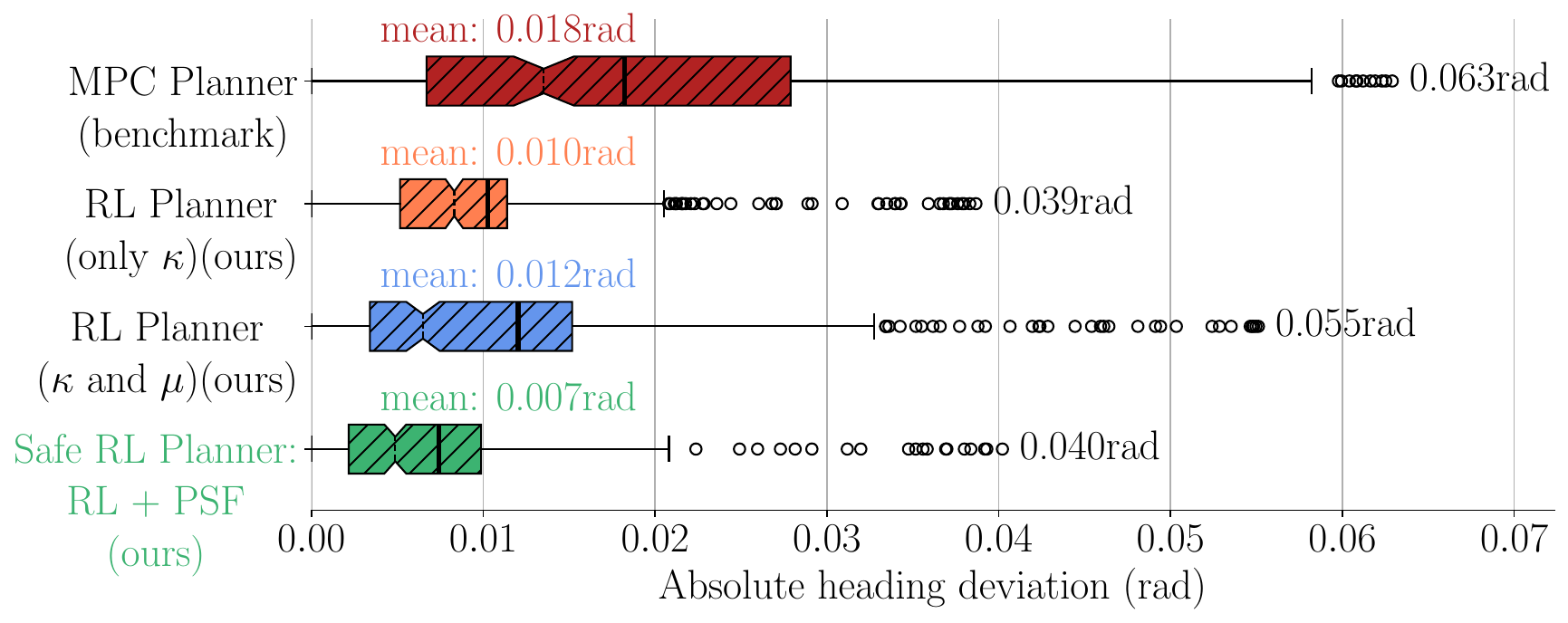}%
		\label{fig:box_zeta}}
        \hfil
	\subfloat[Computational times on an AMD Ryzen 7 5800H CPU (3.20 GHz).]
    {\includegraphics[width=1\linewidth]{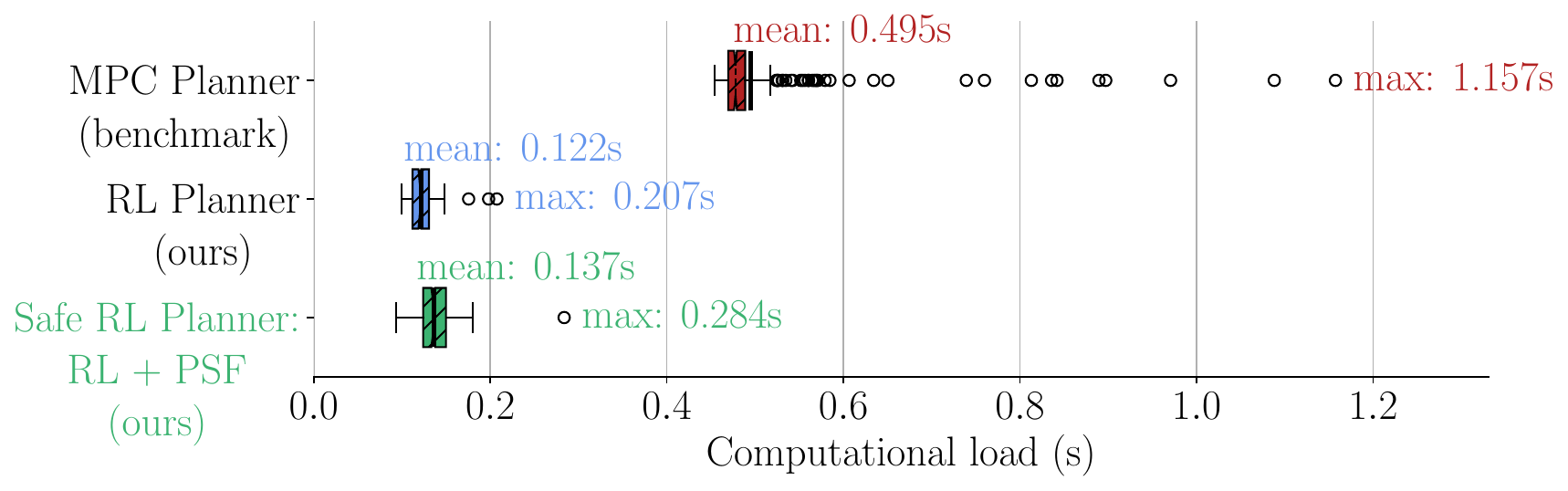}%
		\label{fig:compute}}
	\hfil
	\caption{Box plots of various metrics for different planners during an 18-second drift maneuver. Our proposed Safe RL planner exhibits the smallest lateral and heading deviations, demonstrating superior performance while requiring 3.6 times fewer computational resources on average compared to the MPC benchmark planner \cite{NovelModel2024}.}
	\label{fig:box_plots}
\end{figure}

\begin{figure}[!ht]
	\centering
	\subfloat[Path tracking performance.]
    {\includegraphics[width=0.9\linewidth]{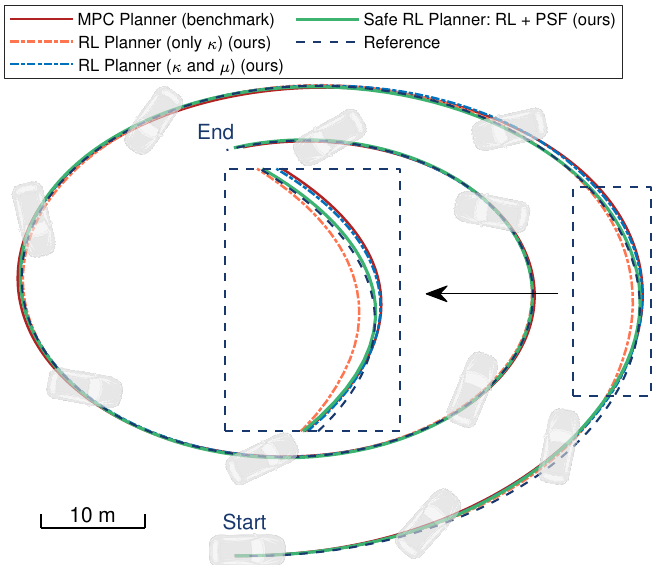}%
	\label{clothoid}}
	\hfil
	\subfloat[Lateral path tracking error.]{\includegraphics[width=0.9\linewidth]{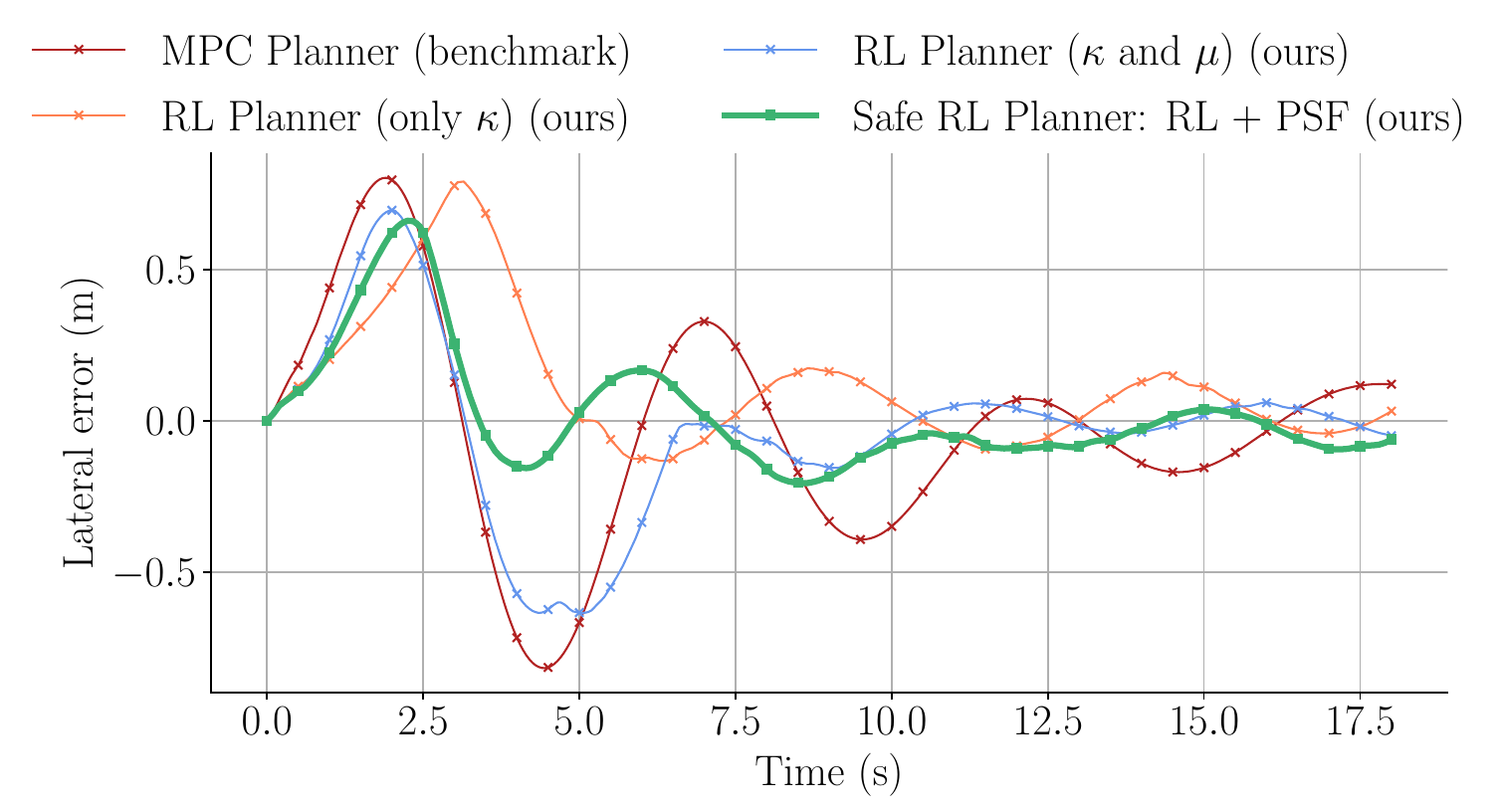}%
		\label{final_e}}
	\hfil
	\caption{Comparisons of path tracking performance among the proposed safe RL planner and benchmark planners. 
    % using an RL agent with the safety filter to plan the tracking curvature and learn modeling errors achieves the best performance.
	} 
	\label{final_tracking}
\end{figure}

We evaluate the impact of learning the reference curvature by comparing the simulation results of two planners: the MPC benchmark \cite{NovelModel2024} and our RL planner that learns only the curvature $\kappa$. Both planners are subject to modeling errors, yet the RL-based approach significantly outperforms the MPC benchmark. As shown in Fig. \ref{fig:box_plots} and \ref{final_tracking}, the RL planner exhibits smaller lateral and heading deviations, resulting in a better tracking stability. 
Compared to the MPC planner, our RL approach improves the mean absolute lateral deviation by \textbf{39.6\%}, the mean absolute heading deviation by \textbf{44.4\%}, and the maximum absolute heading deviation by \textbf{38.1\%}. This demonstrates the RL planner's effectiveness using only the current curvature information, without requiring future trajectory predictions like MPC planners.

\subsection{Learning-Based Correction of Model Inaccuracies}

We compare two RL planners: one learning only the reference curvature \(\kappa\) and another jointly learning \(\kappa\) and the road friction coefficient \(\mu\), both without the Predictive Safety Filter (PSF). Learning \(\mu\) refines the drift equilibrium calculations for the MPDC, enhancing drift control. As shown in Table \ref{tab:drifting}, the RL planner learning both \(\kappa\) and \(\mu\) reduces the drift state's RMSE by {11.5\%} compared to the \(\kappa\)-only planner, and by {14.8\%} compared to the MPC planner. 

While improving the drift accuracy slightly affects the path tracking performance, the trade-off remains minimal, as shown in Fig. \ref{fig:box_plots}. Moreover, training with \(\mu\) yields higher average rewards than training with only \(\kappa\), highlighting the importance of addressing model inaccuracies (Fig. \ref{fig:training}).

\begin{figure*}[ht]
\centering
\subfloat[Episode 500 (Early Termination)] {\includegraphics[width=0.33\linewidth]{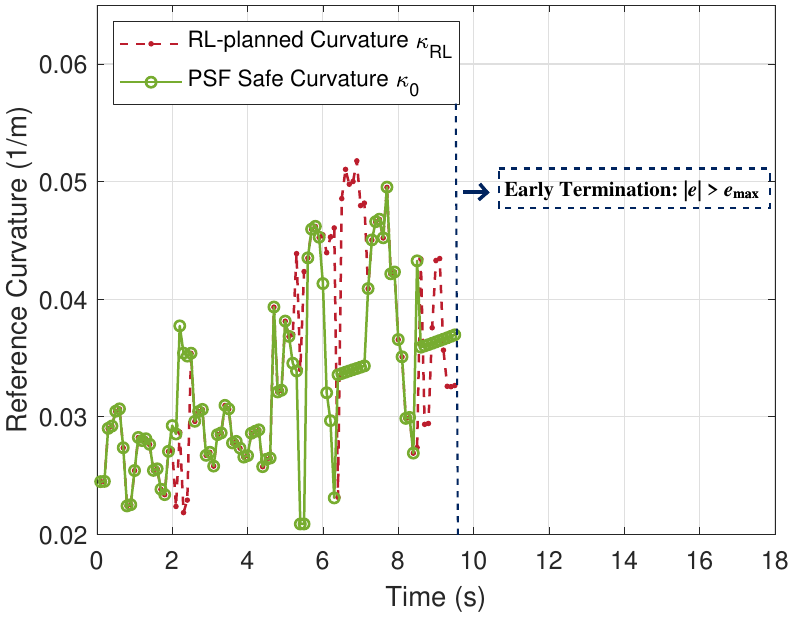}%
	\label{episode_500}}	
\subfloat[Episode 1000] {\includegraphics[width=0.33\linewidth]{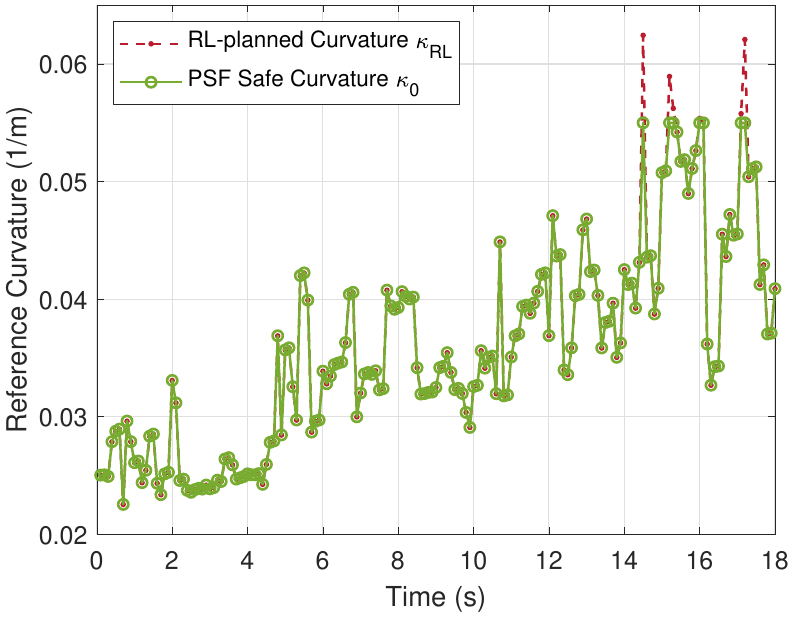}%
	\label{episode_1000}}
\subfloat[Episode 3000] {\includegraphics[width=0.33\linewidth]{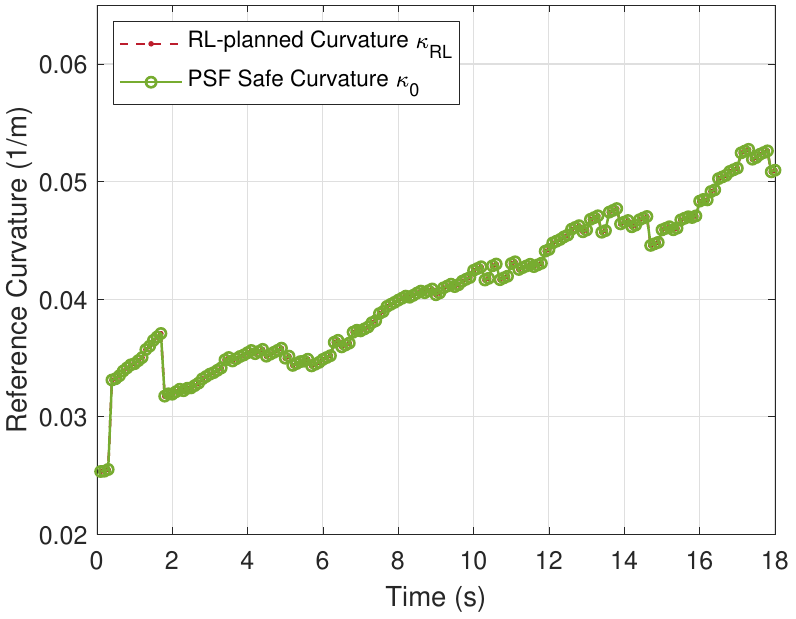}%
	\label{episode_3000}}
\caption{Comparison of RL-planned curvature $\kappa_{\mathrm{RL}}$ and PSF-adjusted safe curvature $\kappa_0$ in specific episodes of the Safe RL planner: As training progresses, the safety filter's adjustments to the RL-planned curvature diminish, reflecting the agent's improved ability to generate safe actions autonomously.}
\label{episode}
\end{figure*}
\begin{table}[ht]
\scriptsize
\renewcommand{\arraystretch}{1.1}
\caption{Drift performance with different motion planners.
% learning the modeling errors can improve the drifting performance.
\label{tab:drifting}}
\centering
\begin{threeparttable}
    \resizebox{\linewidth}{!}{
	\begin{tabular}	{l|ccc} 
		\hline
		\textbf{Planner} &  RMSE $V$ (m/s) & RMSE $\beta$ (rad) & RMSE $r$ (rad/s) \\
		\hline
		MPC Planner (benchmark) & 0.54 & 0.20 & 0.20  \\
		RL Planner (only $\kappa$)  & 0.52 & 0.21 & 0.17  \\		
		RL Planner ($\kappa$ and $\mu$) & \textbf{0.46} & 0.20 & \textbf{0.16} \\
		Safe RL Planner ($\kappa$ and $\mu$ + PSF)  & 0.48 & \textbf{0.19} & 0.17  \\
		\hline
	\end{tabular}
    }
\end{threeparttable}
\end{table}

\subsection{Impact of the Predictive Safety Filter}

The Predictive Safety Filter (PSF) significantly improves the performance of our RL planner, by correcting unsafe actions and preventing large lateral errors. 

Quantitatively (Figs. \ref{fig:box_plots} and \ref{final_tracking}), Safe RL with PSF improves over RL without PSF by \textbf{41.7\%} in mean absolute heading error, \textbf{27.3\%} in max heading deviation, and \textbf{26.0\%} in mean lateral deviation. Compared to the MPC planner, Safe RL shows \textbf{61.1\%} lower mean heading error, \textbf{36.5\%} lower max heading deviation, \textbf{49.6\%} lower mean lateral deviation, and \textbf{19.2\%} lower max lateral deviation.

\subsection{Safety Filter’s Influence During RL Training}
As shown in Fig. \ref{fig:training}, integrating the PSF accelerates training convergence and increases average rewards compared to the RL planner without the PSF. 

Fig. \ref{episode} illustrates the interaction between the RL planner's output $\kappa_{\text{RL}}$ and the PSF's refined reference curvature $\kappa_0$ across three training episodes. 
In the early training stages (Fig. \ref{episode_500}), the PSF heavily adjusts $\kappa_{\text{RL}}$ to enforce safe driving, yet large lateral errors still accumulate, ultimately triggering the termination condition and ending the episode prematurely. 
As training progresses (Fig. \ref{episode_1000}), the RL agent begins producing safer actions, with the PSF making fewer corrections. This results in longer driving episodes, as the agent gradually learns to balance exploration and safety. 
By later stages (Fig. \ref{episode_3000}), the RL agent outputs smoother reference curvatures, reducing the need for PSF intervention. While the safety filter ensures future driving satisfies physical constraints, the RL agent's learned policy further refines its output curvature for more stable and improved vehicle performance.
\subsection{Computation Time}
Fig. \ref{fig:compute} reports the computational times for the MPC, RL, and Safe RL planners. On average, the Safe RL with PSF is \textbf{3.6 times faster} (\textbf{72.3\% improvement}) than the MPC planner and 4.1 times faster (75.5\% improvement) in terms of maximum computational times.

The RL planner training takes 16h 43m, while the Safe RL planner requires 19h 15m. Although MPC planners do not need pre-training, their online optimization steps result in higher inference times, making them less suitable for real-time applications compared to RL planners. Although the safety filter increases the computational training time, the Safe RL planner achieves comparable performance with fewer training episodes (Fig. \ref{fig:training}). For time-critical scenarios, the safety filter could be omitted after sufficient training to reduce computation time.

Finally, the low-level MPDC tracking controller has an average solving time of 0.048 s, for all planners.

\subsection{Generalization Capability}
To assess generalization, we test a trained RL agent on three new unseen tracks with varying curvatures, and compare it to the MPC benchmark planner (Table \ref{generalization}). For a fair comparison, we keep the MPC parameters (e.g., prediction horizon, weights) fixed across all tracks. While both planners exhibit performance degradation on new tracks, the Safe RL planner consistently outperforms the MPC, demonstrating its ability to handle unpredictable environmental changes and its potential for real-world applications.

\begin{table}[!ht]
\caption{Generalization in unseen test tracks: safe RL versus MPC. \label{generalization}}
\centering
\begin{threeparttable}
    \resizebox{\linewidth}{!}{
	\begin{tabular}{ @{\extracolsep{\fill}} c |cc|c  c }        
		\hline
		\multirow{2}{*}{\textbf{Track}}&\multicolumn{2}{ c|}{\textbf{Curvature} (m$^{-1}$)} & \multicolumn{1}{c}{\textbf{MPC Planner}} & \multicolumn{1}{c}{\textbf{Safe RL Planner}} \\	
		%			\cline{2-3} \cline{4-5} \cline{7-8}
		& Start &End  & RMSE $e$ (m) & RMSE $e$ (m) \\			
		\hline
		Training & 1/40 & 1/20 & 0.36  & 0.13 \\
		\hline
		Test Track 1  & 1/45 & 1/20 & 0.49  & 0.35  \\
		Test Track 2  & 1/40 & 1/25 & 0.37  & 0.26  \\
		Test Track 3  & 1/45 & 1/20 & 0.53  & 0.39   \\
		\hline
	\end{tabular}
    }
\end{threeparttable}
\end{table}

\section{Conclusion} \label{sec:concl}

In this work, we introduced a Safe Reinforcement Learning (RL)-based motion planning strategy that optimizes vehicle drift states by simultaneously learning the reference curvature and road friction coefficient in real-time. To ensure safety and feasibility, we incorporated a Predictive Safety Filter (PSF) that dynamically adjusts the RL agent's actions, effectively preventing unsafe states and improving both training efficiency and inference performance. Our simulation results demonstrate that the Safe RL planner significantly outperforms state-of-the-art MPC planners, achieving up to {61.1\%} reduction in mean heading error and {49.6\%} reduction in mean lateral deviation, while requiring {3.6 times} less computational load. Moving forward, we aim to extend this approach to safe online learning on the F1Tenth platform, targeting real-world, safety-critical scenarios with stringent computational constraints.

\bibliographystyle{IEEEtran}
\bibliography{conf_reference}

\begin{thebibliography}{10}
\providecommand{\url}[1]{#1}
\csname url@samestyle\endcsname
\providecommand{\newblock}{\relax}
\providecommand{\bibinfo}[2]{#2}
\providecommand{\BIBentrySTDinterwordspacing}{\spaceskip=0pt\relax}
\providecommand{\BIBentryALTinterwordstretchfactor}{4}
\providecommand{\BIBentryALTinterwordspacing}{\spaceskip=\fontdimen2\font plus
\BIBentryALTinterwordstretchfactor\fontdimen3\font minus
  \fontdimen4\font\relax}
\providecommand{\BIBforeignlanguage}[2]{{%
\expandafter\ifx\csname l@#1\endcsname\relax
\typeout{** WARNING: IEEEtran.bst: No hyphenation pattern has been}%
\typeout{** loaded for the language `#1'. Using the pattern for}%
\typeout{** the default language instead.}%
\else
\language=\csname l@#1\endcsname
\fi
#2}}
\providecommand{\BIBdecl}{\relax}
\BIBdecl

\bibitem{AutonomousVehicles2022a}
J.~Betz, H.~Zheng, A.~Liniger, U.~Rosolia, P.~Karle, M.~Behl, V.~Krovi, and
  R.~Mangharam, ``Autonomous vehicles on the edge: A survey on autonomous
  vehicle racing,'' \emph{IEEE Open Journal of Intelligent Transportation
  Systems}, vol.~3, pp. 458--488, 2022.

\bibitem{Piccinini2024}
M.~Piccinini, S.~Taddei, E.~Pagot, E.~Bertolazzi, and F.~Biral, ``How optimal
  is the minimum-time manoeuvre of an artificial race driver?'' \emph{Vehicle
  System Dynamics}, vol.~0, no.~0, pp. 1--28, 2024.

\bibitem{SimultaneousStabilization2016}
J.~Y. Goh and J.~C. Gerdes, ``Simultaneous stabilization and tracking of basic
  automobile drifting trajectories,'' in \emph{2016 IEEE Intelligent Vehicles
  Symposium (IV)}, Jun. 2016, pp. 597--602.

\bibitem{DynamicDrifting2023}
G.~Chen, X.~Zhao, Z.~Gao, and M.~Hua, ``Dynamic drifting control for general
  path tracking of autonomous vehicles,'' \emph{IEEE Transactions on
  Intelligent Vehicles}, vol.~8, no.~3, pp. 2527--2537, Mar. 2023.

\bibitem{ModelingControl2024a}
T.~P. Weber and J.~C. Gerdes, ``Modeling and control for dynamic drifting
  trajectories,'' \emph{IEEE Transactions on Intelligent Vehicles}, vol.~9,
  no.~2, pp. 3731--3741, Feb. 2024.

\bibitem{AutonomousDrifting2024a}
B.~Lenzo, T.~Goel, and J.~Christian~Gerdes, ``Autonomous drifting using torque
  vectoring: Innovating active safety,'' \emph{IEEE Transactions on Intelligent
  Transportation Systems}, vol.~25, no.~11, pp. 17\,931--17\,939, Nov. 2024.

\bibitem{TeachingVehicle2018}
M.~Acosta and S.~Kanarachos, ``Teaching a vehicle to autonomously drift: A
  data-based approach using neural networks,'' \emph{Knowledge-Based Systems},
  vol. 153, pp. 12--28, Aug. 2018.

\bibitem{HighSpeedAutonomous2020a}
P.~Cai, X.~Mei, L.~Tai, Y.~Sun, and M.~Liu, ``High-speed autonomous drifting
  with deep reinforcement learning,'' \emph{IEEE Robotics and Automation
  Letters}, vol.~5, no.~2, pp. 1247--1254, Apr. 2020.

\bibitem{DeepDrifting2022}
F.~Domberg, C.~C. Wembers, H.~Patel, and G.~Schildbach, ``Deep drifting:
  Autonomous drifting of arbitrary trajectories using deep reinforcement
  learning,'' in \emph{2022 International Conference on Robotics and Automation
  (ICRA)}, May 2022, pp. 7753--7759.

\bibitem{AutonomousDrifting2016b}
M.~Cutler and J.~P. How, ``Autonomous drifting using simulation-aided
  reinforcement learning,'' in \emph{2016 IEEE International Conference on
  Robotics and Automation (ICRA)}, May 2016, pp. 5442--5448.

\bibitem{djeumou2024one}
F.~Djeumou, T.~J. Lew, N.~DING, M.~Thompson, M.~Suminaka, M.~Greiff, and
  J.~Subosits, ``One model to drift them all: Physics-informed conditional
  diffusion model for driving at the limits,'' in \emph{8th Annual Conference
  on Robot Learning}, 2024.

\bibitem{LearningBasedMPC2022}
X.~Zhou, C.~Hu, R.~Duo, H.~Xiong, Y.~Qi, Z.~Zhang, H.~Su, and L.~Xie,
  ``Learning-based mpc controller for drift control of autonomous vehicles,''
  in \emph{2022 IEEE 25th International Conference on Intelligent
  Transportation Systems (ITSC)}.\hskip 1em plus 0.5em minus 0.4em\relax Macau,
  China: IEEE, Oct. 2022, pp. 322--328.

\bibitem{AutonomousDriving2022a}
X.~Hou, J.~Zhang, C.~He, Y.~Ji, J.~Zhang, and J.~Han, ``Autonomous driving at
  the handling limit using residual reinforcement learning,'' \emph{Advanced
  Engineering Informatics}, vol.~54, p. 101754, Oct. 2022.

\bibitem{HarmonizedApproach2024a}
S.~Zhao, J.~Zhang, X.~He, C.~He, X.~Hou, H.~Huang, and J.~Han, ``A harmonized
  approach: Beyond-the-limit control for autonomous vehicles balancing
  performance and safety in unpredictable environments,'' \emph{IEEE
  Transactions on Intelligent Transportation Systems}, pp. 1--14, 2024.

\bibitem{NovelModel2024}
C.~Hu, L.~Xie, Z.~Zhang, and H.~Xiong, ``A novel model predictive controller
  for the drifting vehicle to track a circular trajectory,'' \emph{Vehicle
  System Dynamics}, vol.~0, no.~0, pp. 1--30, 2024.

\bibitem{HighSpeedCornering2024}
S.~Zhao, J.~Zhang, N.~Masoud, Y.~Jiang, H.~Huang, and T.~Liu, ``High-speed
  cornering control and real-vehicle deployment for autonomous electric
  vehicles,'' Nov. 2024.

\bibitem{gu2024review}
S.~Gu, L.~Yang, Y.~Du, G.~Chen, F.~Walter, J.~Wang, and A.~Knoll, ``A review of
  safe reinforcement learning: Methods, theories and applications,'' \emph{IEEE
  Transactions on Pattern Analysis and Machine Intelligence}, 2024.

\bibitem{selim2022safe}
M.~Selim, A.~Alanwar, M.~W. El-Kharashi, H.~M. Abbas, and K.~H. Johansson,
  ``Safe reinforcement learning using data-driven predictive control,'' in
  \emph{2022 5th International Conference on Communications, Signal Processing,
  and their Applications (ICCSPA)}.\hskip 1em plus 0.5em minus 0.4em\relax
  IEEE, 2022, pp. 1--6.

\bibitem{pmlr-v162-sootla22a}
A.~Sootla, A.~I. Cowen-Rivers, T.~Jafferjee, Z.~Wang, D.~H. Mguni, J.~Wang, and
  H.~Ammar, ``Saute {RL}: Almost surely safe reinforcement learning using state
  augmentation,'' in \emph{Proceedings of the 39th International Conference on
  Machine Learning}, vol. 162.\hskip 1em plus 0.5em minus 0.4em\relax PMLR,
  17--23 Jul 2022, pp. 20\,423--20\,443.

\bibitem{yang2023safe}
W.-C. Yang, G.~Marra, G.~Rens, and L.~De~Raedt, ``Safe reinforcement learning
  via probabilistic logic shields,'' \emph{arXiv preprint arXiv:2303.03226},
  2023.

\bibitem{Safe-RL-Zarrouki}
B.~Zarrouki, M.~Spanakakis, and J.~Betz, ``A safe reinforcement learning driven
  weights-varying model predictive control for autonomous vehicle motion
  control,'' in \emph{2024 IEEE Intelligent Vehicles Symposium (IV)}, 2024, pp.
  1401--1408.

\bibitem{hsu2023safety}
K.-C. Hsu, H.~Hu, and J.~F. Fisac, ``The safety filter: A unified view of
  safety-critical control in autonomous systems,'' \emph{Annual Review of
  Control, Robotics, and Autonomous Systems}, vol.~7.

\bibitem{pmlr-v242-lavanakul24a}
W.~Lavanakul, J.~Choi, K.~Sreenath, and C.~Tomlin, ``Safety filters for
  black-box dynamical systems by learning discriminating hyperplanes,'' in
  \emph{Proceedings of the 6th Annual Learning for Dynamics amp; Control
  Conference}, ser. Proceedings of Machine Learning Research, vol. 242.\hskip
  1em plus 0.5em minus 0.4em\relax PMLR, 15--17 Jul 2024, pp. 1278--1291.

\bibitem{Brunke}
F.~P. Bejarano, L.~Brunke, and A.~P. Schoellig, ``Safety filtering while
  training: Improving the performance and sample efficiency of reinforcement
  learning agents,'' \emph{IEEE Robotics and Automation Letters}, vol.~10,
  no.~1, pp. 788--795, 2025.

\bibitem{WABERSICH2021109597}
K.~P. Wabersich and M.~N. Zeilinger, ``A predictive safety filter for
  learning-based control of constrained nonlinear dynamical systems,''
  \emph{Automatica}, vol. 129, p. 109597, 2021.

\bibitem{pmlr-v211-leeman23a}
A.~Leeman, J.~K\"ohler, S.~Bennani, and M.~Zeilinger, ``Predictive safety
  filter using system level synthesis,'' in \emph{Proceedings of The 5th Annual
  Learning for Dynamics and Control Conference}, vol. 211.\hskip 1em plus 0.5em
  minus 0.4em\relax PMLR, 15--16 Jun 2023, pp. 1180--1192.

\bibitem{TyreModelling1987}
E.~Bakker, L.~Nyborg, and H.~B. Pacejka, ``Tyre modelling for use in vehicle
  dynamics studies,'' \emph{SAE Transactions}, pp. 190--204, 1987.

\bibitem{AutomatedVehicle2019}
J.~Y. Goh, T.~Goel, and J.~Christian~Gerdes, ``Toward automated vehicle control
  beyond the stability limits: Drifting along a general path,'' \emph{Journal
  of Dynamic Systems, Measurement, and Control}, vol. 142, no. 021004, Nov.
  2019.

\bibitem{ControllerFramework2014a}
R.~Y. Hindiyeh and J.~Christian~Gerdes, ``A controller framework for autonomous
  drifting: Design, stability, and experimental validation,'' \emph{Journal of
  Dynamic Systems, Measurement, and Control}, vol. 136, no. 051015, Jul. 2014.

\bibitem{LearningBasedHierarchical2024}
B.~Zhou, C.~Hu, Y.~Shi, X.~Hu, L.~Xie, and H.~Su, ``Learning-based hierarchical
  model predictive control for drift vehicles,'' in \emph{2024 American Control
  Conference (ACC)}, Jul. 2024, pp. 3524--3530.

\bibitem{ContinuousControl2019}
T.~P. Lillicrap, J.~J. Hunt, A.~Pritzel, N.~Heess, T.~Erez, Y.~Tassa,
  D.~Silver, and D.~Wierstra, ``Continuous control with deep reinforcement
  learning,'' Jul. 2019.

\bibitem{PredictiveSafety2021a}
B.~Tearle, K.~P. Wabersich, A.~Carron, and M.~N. Zeilinger, ``A predictive
  safety filter for learning-based racing control,'' \emph{IEEE Robotics and
  Automation Letters}, vol.~6, no.~4, pp. 7635--7642, Oct. 2021.

\bibitem{ZHOU2025104941}
B.~Zhou, C.~Hu, J.~Zeng, Z.~Li, J.~Betz, L.~Xie, and H.~Su, ``Adaptive
  learning-based model predictive control strategy for drift vehicles,''
  \emph{Robotics and Autonomous Systems}, p. 104941, 2025.

\end{thebibliography}

\end{document}